\documentclass[11pt,a4paper]{article}
\usepackage[hyperref]{emnlp2020}
\usepackage{times}
\usepackage{latexsym}
\usepackage{graphicx}
\usepackage{booktabs}
\usepackage{amsmath}

\usepackage{enumitem}
\usepackage{textcomp}
\usepackage{caption}
\usepackage{subcaption}
\usepackage{comment}

\usepackage{microtype}

\usepackage[nameinlink]{cleveref}
\crefformat{subsection}{\S#2#1#3}
\crefformat{section}{\S#2#1#3}
\Crefname{equation}{Eq.}{Eqs.}
\Crefname{figure}{Fig.}{Figs.}
\crefformat{subsubsection}{\S#2#1#3}
\newcommand{\wrt}{\emph{w.r.t.}\ }
\newcommand{\ie}{{\em i.e.,}\ }
\aclfinalcopy 

\usepackage{bbm}

\usepackage{amsmath,amsfonts,bm}

\definecolor{mypink3}{cmyk}{0, 0.7808, 0.4429, 0.1412}

\definecolor{mypink2}{RGB}{0, 100, 250}

\def\eqref#1{equation~\ref{#1}}

\def\1{\bm{1}}

\def\vt{{\bm{t}}}

\def\vv{{\bm{v}}}
\def\vw{{\bm{w}}}

\def\mT{{\bm{T}}}
\def\mU{{\bm{U}}}

\def\mW{{\bm{W}}}

\DeclareMathAlphabet{\mathsfit}{\encodingdefault}{\sfdefault}{m}{sl}
\SetMathAlphabet{\mathsfit}{bold}{\encodingdefault}{\sfdefault}{bx}{n}

\def\gD{{\mathcal{D}}}

\def\gL{{\mathcal{L}}}

\def\gR{{\mathcal{R}}}

\newcommand{\softmax}{\mathrm{softmax}}
\newcommand{\sigmoid}{\mathrm{sigmoid}}

\DeclareMathOperator*{\argmax}{arg\,max}

\title{Response Selection for Multi-Party Conversations with\\ Dynamic Topic Tracking}

\author{Weishi Wang$^\S{}^{\ddagger}$, Shafiq Joty$^\S{}^{\ddagger}$, Steven C.H. Hoi$^\S$\\
  $^\S$Salesforce Research Asia\\
  $^{\ddagger}$Nanyang Technological University, Singapore \\
  $^\S$\texttt{\{weishi.wang,sjoty,shoi\}@salesforce.com} \\ 
}
\date{}

\begin{document}
\maketitle
\begin{abstract}

While participants in a multi-party multi-turn conversation simultaneously engage in multiple conversation topics, existing response selection methods are developed mainly focusing on a two-party single-conversation scenario. Hence, the prolongation and transition of conversation topics are ignored by current methods. In this work, we frame response selection as a dynamic topic tracking task to match the topic between the response and relevant conversation context. With this new formulation, we propose a novel multi-task learning framework that supports efficient encoding through large pretrained models with only two utterances at once to perform dynamic topic disentanglement and response selection. We also propose Topic-BERT an essential pretraining step to embed topic information into BERT with self-supervised learning. Experimental results on the DSTC-8 Ubuntu IRC dataset show state-of-the-art results in response selection and topic disentanglement tasks outperforming existing methods by a good margin.\footnote{Code is available at \url{https://github.com/salesforce/TopicBERT}.}




 
 

\end{abstract}

\section{Introduction}

In recent years, with the influx of deep learning methods in natural language processing (NLP), there has been a lot of interests in building effective task-oriented dialogue systems that can assist people in real-world business such as booking tickets, ordering food and solving technical issues \cite{Trung-2006}. \textit{Retrieval-based} response generation that selects a suitable response from a pool of candidates (pre-existing human responses) has become a popular approach to framing dialog. Compared to the \textit{generation-based} systems that generate novel utterances \cite{serban2016generative}, retrieval-based systems produce fluent, grammatical and informative responses \cite{weston2018retrieve,henderson2019convert}. Also compared to the traditional modular approach, it does not rely on dedicated modules for language understanding, dialog management, and generation, thus simplifying the system design. Due to these reasons, retrieval-based systems have been widely adopted in commercial dialogue systems \cite{GaoconvAI,gunasekara2019dstc7}.

\begin{figure}[t!]
    \centering
    \includegraphics[scale=.42]{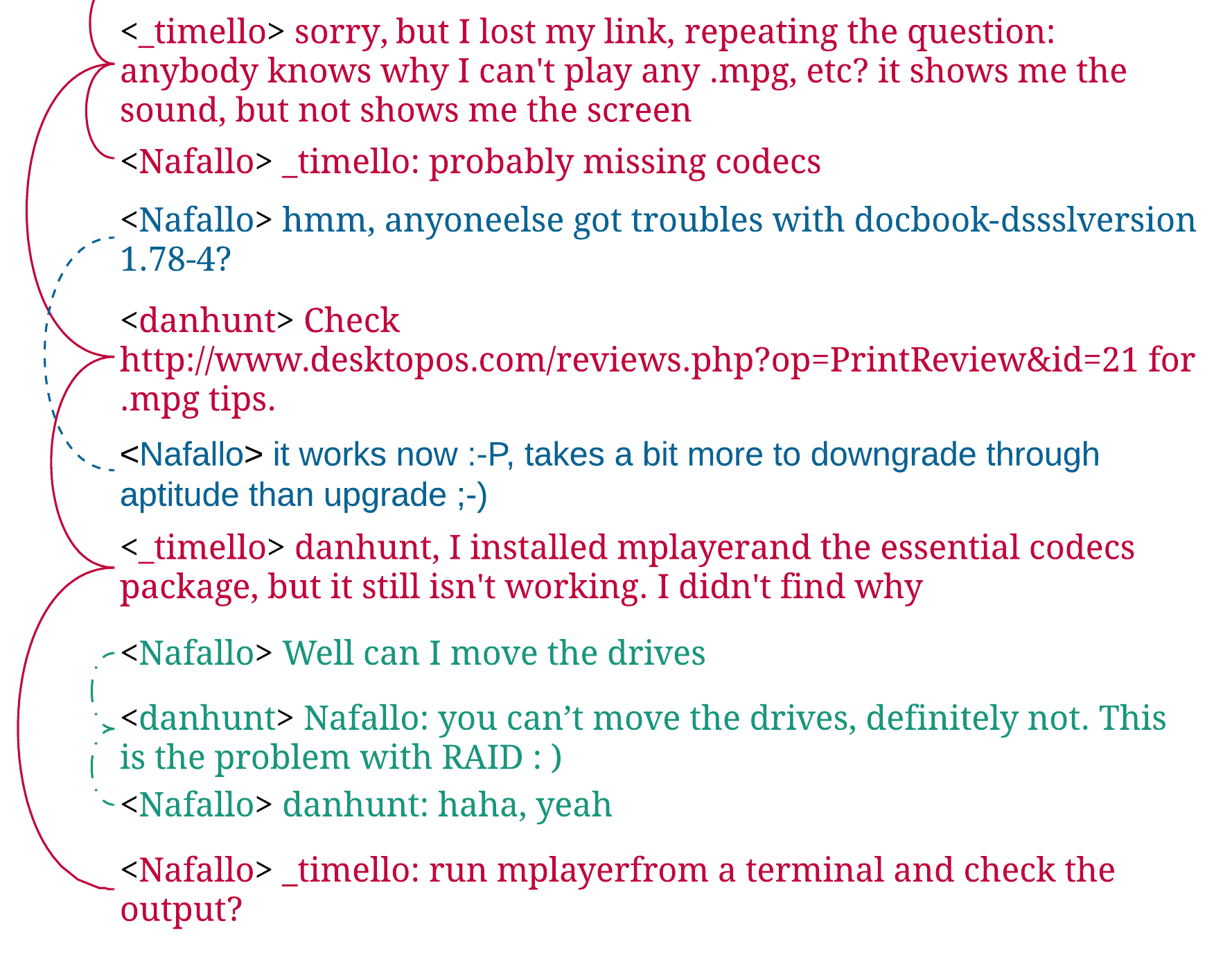}
    \caption{A (truncated) multi-party conversation from Ubuntu IRC log. Curved arrows show the `reply-to' links between utterances. We use different colors to represent different conversation topic clusters.} 
    \label{fig:dialogue example}
\end{figure}

Initially, researchers considered response selection in single-turn conversations, where only the last input utterance is considered as the context query \cite{10.1145/2911451.2911542}. More recent work deals with multi-turn context, which shows improvements over the single-turn context \cite{lowe2015ubuntu,Lowe2017TrainingED,zhou2016multi,ESIM2,IMN,DAM}. These methods typically aim to encode the context and the candidate responses in a joint semantic space by capturing short and long range dependencies, and then retrieve the most relevant response by matching the query representation against each candidate's representation through attentions.




However, most of these works are limited to only two-party conversations. As dialogue research progresses, it is necessary to study the more generic multi-party multi-turn scenario, which has become very common (e.g., Slack, Whatsapp) with the advent of Internet and mobile devices, and posits a unique set of challenges for the dialog models \cite{kim2019eighth,acl19disentangle}.

Multiple ongoing conversations seem to occur more naturally in multi-party conversations. For example, consider the conversation excerpt in Figure \ref{fig:dialogue example} among three participants, taken from the Ubuntu IRC corpus. There are three ongoing conversation topics as highlighted by different color, and the participants contribute to multiple topics simultaneously (e.g., \texttt{Nafallo} participates in three and \texttt{danhunt}  participates in two). An effective response selection method should model such complex conversational topic dynamics in the context, for which existing methods are deficient. In particular, a proper response should match with its context in terms of the same conversation topic, while ignoring other non-relevant topics. 



To address the aforementioned challenges in multi-party multi-turn dialog, we frame response selection as a dynamic topic tracking task with the intuition that the topic should remain the same as we go from the context to the response. Our formulation is also supported by the Segmented Discourse Representation Theory (SDRT) of  conversations \cite{asher:lascarides:2003}. Based on this new formulation, we propose a novel architecture that can incorporate other related dialog tasks such as conversation disentanglement, enabling multi-task learning in a unified framework.

Crucially, our formulation of the task needs to encode only two utterances at a time, thus allowing efficient encoding via large pretrained models like BERT \cite{BERT}. Furthermore, it facilitates pretraining of BERT-like models on topic related sentence pairs to incorporate topic relevance in pretraining, which can be done on large dialog corpora with self-supervised objectives, requiring no manual topic annotations, and can benefit not only response selection but also other dialog tasks. In summary, our contributions are:

\begin{itemize}[leftmargin=*,itemsep=0.1pt]  

\item A new formulation of the response selection task with an efficient multi-task learning framework for dynamic topic tracking, which supports efficient encoding with only two utterances at once.

\item Incorporate topic prediction and topic disentanglement as auxiliary tasks within the framework. Based on the similarity of these three tasks, the objective is to match topic (topic prediction) between context utterance and response and track response's topic (topic disentanglement) across contexts to select an appropriate response.

\item Propose Topic-BERT as a pretraining step to embed topic information into BERT, and use a self-supervised approach to generate topic sentence pairs from existing dialogue datasets. The incorporated topic information is shown to be a key step to our topic tracking framework.

\item Apply topic attention by using topic embedding as query to obtain utterance-level embeddings for topic prediction. Then self-attention was applied to capture the contextual topic vectors for response selection and topic disentanglement. 

\item Evaluate the proposed models on the DSTC-8 Ubuntu IRC dataset \cite{kim2019eighth}, and show state-of-the-art results in both response selection and topic disentanglement outperforming the existing methods by a good margin.

\end{itemize}

\section{Related Work}


\subsection{Response Selection} 

A dual encoder framework was proposed to match the context and response \cite{lowe2015ubuntu}, and the long short-term memory (LSTM) was utilized to learn the long and short term dependencies among tokens. Beyond tokens, the sentence view matching was introduced by applying a hierarchical recurrent neural network to model sentence level relationships \cite{zhou-etal-2016-multi}. However, context utterances and response are encoded separately without interaction; thus the semantics extracted from context are not based on the response. Recent approaches such as Sequential Matching Netowrk (SMN) \cite{SMN} leverage the contextual information by matching each contextual utterance with response and the multi channel Convolutional Neural Network (CNN) was proposed to generate multiple levels of granularity of matched segment.

These hierarchy-based methods use LSTM to encode the text, which is not cost effective to capture multi-grained segment representation \cite{lowe2015ubuntu,zhou-etal-2016-multi,SMN}. A particular work on sequence-based method stand out in DSTC-7; Enhanced Sequential Inference Model (ESIM) \cite{ESIM} achieves the state-of-the-art performance in DSTC-7 by taking advantage of inter-sentence matching \cite{ESIM1,ESIM2}. It converts multi-turn dialogue setting to natural language inference setting. In addition, transformer-based approach Deep Attention Matching (DAM) solve response selection problem by attention mechanism \cite{DAM}. It utilizes utterance self-attention and context-to-response cross attention to leverage the hidden representation at multi-grained level. Similar to DAM, Multi-hop Selector Network (MSN) was proposed by \citet{yuan2019multi} to fuse and select relevant context utterances and match it with the response utterance. In addition, \citet{tao2019one} model the relationship between a context utterance and the response in multiple levels.

Compared to LSTM-based approaches, methods based on transformers \cite{transformer}  present a promising performance in both accuracy and efficiency \cite{BERT-SM}. \citet{BERT} proposed BERT, a  transformer-based large-scale pretrained language model, which achieves state-of-the-art performance in different NLP tasks. BERT is also a good match to response selection problem as shown by \citet{vig2019comparison}. Our Topic-BERT is initialised with BERT$_{base}$ and posttrained with topic related sentence pairs.

\subsection{Hard Context Retrieval} 

The side effect of multi-speaker multi-turn context is crucial; a lot of noise will be introduced in the context utterances. The speaker and addressee information are essential to decide the structure of conversation, thus can also benefit conversational response selection \cite{zhang2017addressee,le2019speaking,hu2019gsn}. 
A hard context retrieval method was proposed by \citet{BERT-enhance} to minimize the context size, while keeping only the utterances whose speaker is the same as the response candidates or referred by the response candidates. 
However, it cannot guarantee clean context with a single topic of conversation. Indeed, topic tracking is necessary along with hard context retrieval. 

\subsection{Conversation Disentanglement}

Traditional statistical learning  based approaches and linguistic features have shown to be effective for conversation disentanglement \cite{mayfield2012hierarchical, du2016discovering}. Recent methods demonstrate that  neural networks could be applied to have a better linguistic representation of the utterances to retrieve relevant conversation. Hand crafted features and pretrained word embeddings are utilized to predict the link-to relationship between utterances \cite{acl19disentangle}. Recently, BERT has been adapted in disentangling task to capture the semantics across utterances \cite{gu2020pre}. Also, a masked transformer has been applied to learn the graphical representation {of utterances based on the reply-to links \cite{zhu2019did}}. \citet{Tao-emnlp-20} apply a pointer network for online disentanglement of conversations.

\section{Task Formulation}
\label{sec:taskform}

Our Topic-BERT framework combines response selection task with two auxiliary tasks, which are topic prediction and topic disentanglement.  

\paragraph{Response Selection} Our primary task is response selection in multi-party multi-turn conversations. Let $\gD_{rs} = \{(c_{i},r_{i,j},y_{i,j})\}_{i=1}^{|\gD_{rs}|}$ is a  response selection dataset, where $j$ is the index of a response candidate for a context $c_i = \{u_{1},u_{2},\ldots, u_{n}\}$ with $n$ utterances. Each utterance $u_{i} = \{s_{i},w_{i,1}, w_{i,2}, \ldots,  w_{i,m}\}$ starts with its speaker $s_i$ and is composed of $m$ words. Similarly, a response $r_{i,j}$ has a speaker $s_{i,j}$ and composed of $n$ words. $y_{i,j} \in \{0,1\}$ represents the relevance label. Our goal is to find the relevance ranking score $f_{\theta_{r}}(c_{i},r_{i,j})$ with model parameters ${\theta_{r}}$.



\paragraph{Topic Prediction} For this (auxiliary) task, we assume a multi-party conversation with a single conversation topic. Let $\gD_{tp} = \{(c_{i},r_i^{+},r_{i,j}^{-})\}_{i=1}^{|\gD_{tp}|}$ is a topic prediction dataset, where $r_i^{+}$ is a positive (same conversation) response and $r_{i,j}^{-}$ is a negative (difference conversation) response for context $c_{i}$. For our training purposes, each utterance pair from the same context constitutes $(c_{i},r_i^{+})$, whereas an utterance pair from different contexts constitutes $(c_{i},r_{i,j}^{-})$. Our goal is to train a binary classifier $g_{\theta_{t}}(c_{i},r_{i}) \in \{0,1\} $ with model parameters ${\theta_{t}}$. 



\paragraph{Topic Disentanglement} In this (auxiliary) task, our goal is to disentangle single conversations from a multi-party conversation based on topics. For a given conversation context $c_i = \{u_{1},u_{2}, \ldots, u_{n}\}$, a set of pairwise ``reply-to'' annotations $\gR = \{(u_c, u_p)_1, \ldots, (u_c, u_p)_{|\gR|} \}$ is given, where $u_p$ is a parent of child $u_c$. Our task is to compute a reply-to score $h_{\theta_d} (u_i, u_j)$ for $j \le i$ that indicates the score for $u_j$ being the parent of $u_i$, with model parameters ${\theta_d}$. The individual conversations can then be constructed by following the reply-to links. Note that an utterance $u_i$ can point to itself, which we call \emph{self-link}. Self-links are either start of a conversation or a system message, and they play a crucial role in identifying the conversation clusters. 
\begin{figure*}[t!]
    \centering
    \includegraphics[scale=.52]{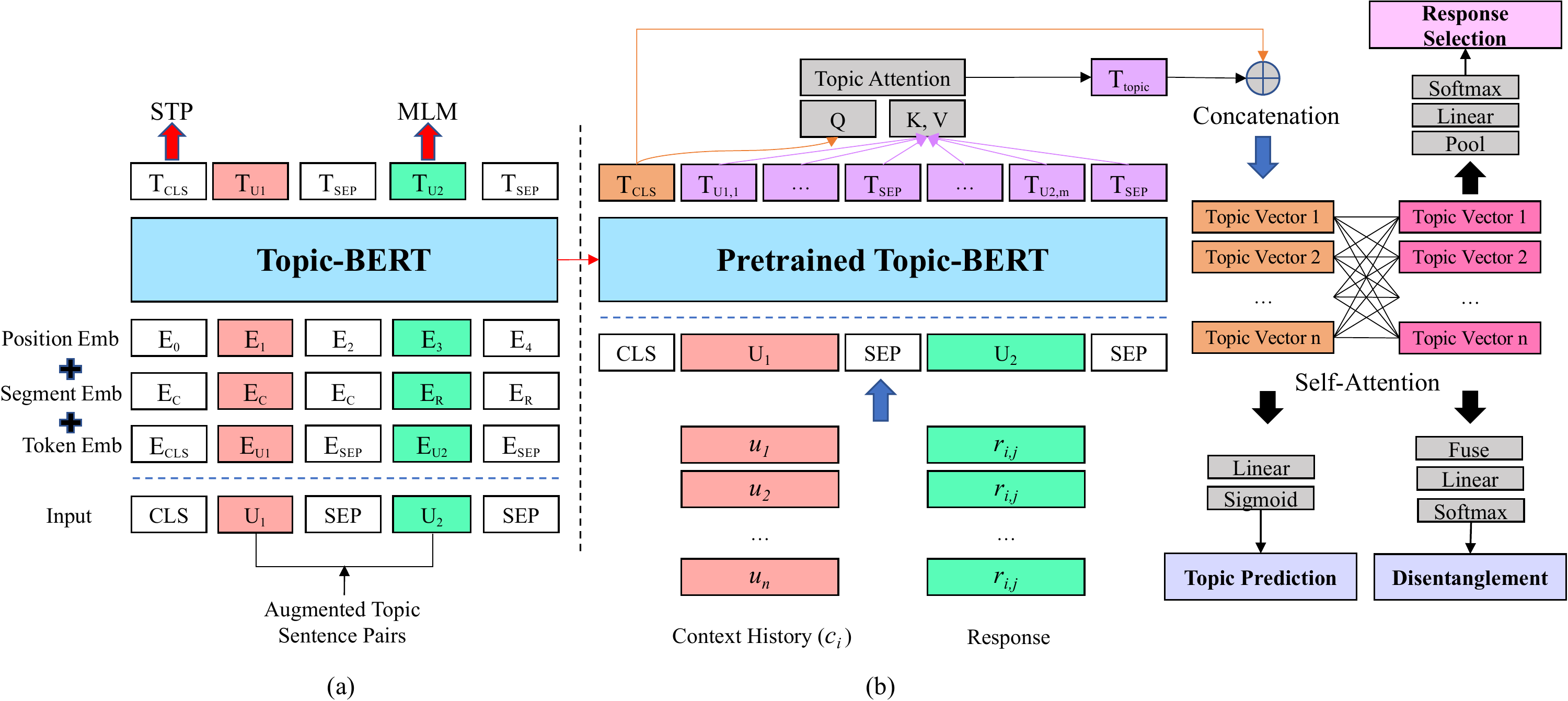}
    \caption{Overview of Topic-BERT architecture. (a) Topic-BERT pretraining with topic sentence pairs to incorporate utterance-utterance topic relationship. (b) Our multi-task framework which uses the pretrained Topic-BERT to enhance topic information in the encoded representations to support three downstream tasks -- response selection as the main task while topic prediction and disentanglement as two auxiliary (optional) tasks.}
\label{fig:model}
\end{figure*}

\section{Our Topic-BERT Framework}
Our framework for response selection aims to track how the conversation topics change from one utterance to another and use it for ranking the candidate responses.  As shown in \Cref{fig:model}, we encode an utterance $u_k$ from the context $c_i = \{u_{1},u_{2}, \ldots, u_{n}\}$ along with a candidate response $r_{i,j}$ using our pretrained Topic-BERT encoder (\cref{Self-supervised Topic-BERT Pretraining}). The contextual token representations in Topic-BERT encode topic relevance between the tokens of $u_k$ and the tokens of $r_{i,j}$, while the \texttt{[CLS]} representation captures utterance-level topic relevance. We use the \texttt{[CLS]} representation as query to attend over the token representations to further enforce topic relevance in the attended topic vector $\vt_{^k}$.

We repeat this encoding process for the $n$ utterances in the context $c_i$ to get $n$ different topic vectors $\mT_j = \{\vt_1, \ldots, \vt_n \}$ that model $r_{i,j}$'s topic relevance to the each of context utterances. These topic representations are then used for the prediction tasks -- topic prediction, disentanglement, and response selection. Response selection is our main task, while the other two tasks are auxiliary and optional. Since our Topic-BERT  encodes two utterances at a time, the encoding process is efficient and can be used to encode larger context. The core component of our framework is the Topic-BERT pretraining as we describe next.

\subsection{Topic-BERT Pretraining}
\label{Self-supervised Topic-BERT Pretraining}

One crucial advantage of our topic-based task formulation is that it allows us to pretrain BERT directly on a very relevant task in a self-supervised way, without requiring any human annotation. In other words, our goal is to pretrain BERT such that it can be used to encode relevant topic information for our task(s). For this, we assume that a single-threaded conversation between {two or more  participants} covers a single topic and the utterance pairs in that thread can be used to pretrain our Topic-BERT with relevant self-supervised objectives.  

To collect such single-threaded conversational data in an opportunistic way, we can simply adopt the heuristics (unsupervised) used by \citet{lowe2015ubuntu} to collect the popular Ubuntu Dialogue Corpus from multi-threaded chatlogs. Alternatively, we can extract two-party conversations from other sources as done in previous work  \cite{henderson2019convert,wu2020tod}. In our experiments, we use the data from DSTC-8 task 1 \cite{kim2019eighth}, which was automatically collected from Ubuntu chat logs. 
This dataset contains detached speaker-visible conversations between two or more participants from the Ubuntu IRC channel. 


To pretrain Topic-BERT, we first initialise it with the pretrained uncased BERT$_{base}$ \cite{BERT}. We treat the training setting similar to our \emph{topic predection} task in \Cref{sec:taskform}. Formally, the pretraining dataset is $\gD_{pr} = \{(u_{i},r_i^{+},r_{i,j}^{-})\}_{i=1}^{|\gD_{pr}|}$, where each utterance pair from the same conversation (including the true response) constitutes a positive pair $(u_{i},r_i^{+})$, and for each such positive pair we randomly sample 4 negative responses ($r_{i,j}^{-}$) from the 100 candidate pool to balance the positive and negative ratio. We (re)train Topic-BERT on $\gD_{pr}$ with two self-supervised objectives as follows.

\paragraph{Masked Language Modeling (MLM)} We follow the same MLM training of the original BERT \cite{BERT} by masking 15\% of the input tokens at random, and replacing the masked word with \texttt{[MASK]} token at 80\% of the time, with a random word at 10\% of the time, and with the original word at 10\% of the time. The MLM objective is only applied to the positive samples.

\paragraph{Same Topic Prediction (STP)}  Each training pair ($(u_{i},r_i^{+})$ or $(u_{i},r_{i,j}^{-})$) is fed into the Topic-BERT as (\texttt{[CLS]}, $[u_1]$, \texttt{[SEP]}, $[u_2]$, \texttt{[SEP]}). Similar to the original BERT's Next Sentence Prediction (NSP) task, the position embedding, segment embedding and token embedding are added together to get input layer token representations. The token representations are then passed through multiple transformer \cite{transformer} encoder layers, where each layer is comprised of a self-attention and a feed-forward sublayer. Different from the original BERT, Topic-BERT uses the \texttt{[CLS]} representation to predict whether the training instance is a positive (same topic) pair or a negative (different topic) pair. Thus, the \texttt{[CLS]} representation encodes topic relationship between the two utterances and will be used as the topic-aware contextual embedding to determine whether the two utterances are matched in topic. 








\subsection{Topic-BERT Multi-Task Framework}

As shown in \Cref{fig:model}(b), the encoded representations from our Topic-BERT are passed through a topic attention layer (\cref{sec:topicattn}) to get the corresponding topic vectors, which are then used for the end tasks.


\subsubsection{Topic Attention Layer} \label{sec:topicattn}

We apply an attention layer to enhance topic information in the encoded vector. We use the Topic-BERT's \texttt{[CLS]} representation $T_{\textsc{cls}}$ as query to attend to the remaining $K$ tokens $\{T_j\}_{j=1}^K$: 
\begin{eqnarray}
\small
\label{eq:Topic Attention1}
\centering
    e_{j} &=& \vv_{a}^{T} \tanh (\mW_{a} T_{\textsc{cls}} + \mU_{a}T_{j}); \\
     a_{j} &=& \frac{\exp({e_{j}})}{\sum_{j=1}^{K}\exp{(e_{i}})} \\
T_{\text{topic}} &=& \sum_{j=1}^{K} a_{j}T_{j}
\end{eqnarray}

where $\vv_a$, $\mW_a$ and $\mU_a$ are trainable parameters. The concatenation of $T_{\text{topic}}$ and $T_{\textsc{cls}}$ constitutes the final topic vector, \ie\ $\vt = [T_{\textsc{cls}};T_{\text{topic}}]$. We repeat this encoding process for the $n$ utterances in the context $c_i = \{u_{1},u_{2}, \ldots, u_{n}\}$ by pairing each with the candidate response $r_{i,j}$ to get $n$ different topic vectors $\mT_j = \{\vt_1, \ldots, \vt_n \}$. $\mT_j$ represents $r_{i,j}$'s topic relevance to the  context utterances, which will be fed to the task-specific layers.









\subsubsection{Topic Prediction} \label{sec:toppred}

Topic prediction is done for each utterance-response pair $(u_k, r_{i, j})$ for all $u_k \in c_i$ to decide whether $u_k$ and $r_{i, j}$ should be in the same topic (\cref{sec:taskform}). The Topic-BERT encoded topic vector corresponding to the $(u_k, r_{i, j})$ pair is $\vt_k \in \mT_j$. We define the binary topic classification model as:

\begin{equation}
\centering
    g_{{\theta_{t}}}(u_k, r_{i,j}) = \sigmoid( \vw_{p}^T \vt_k)
\end{equation}
where $\vw_p$ is the task-specific parameter. We use a binary cross entropy loss computed as:
\begin{equation}
\begin{split}
\label{eq:topic prediction loss}
    \gL_{\text{topic}} = - y \log (g_{{\theta_{t}}}) - (1-y) \log (1-g_{{\theta_{t}}})
\end{split}
\end{equation}
where $y \in \{0,1\}$ is the ground truth indicating same or different topic. Note that topic prediction is an auxiliary task intended to help our main task of response selection, as we describe next.   


\subsubsection{Response Selection}


In response selection, our goal is to measure relevance of a candidate response $r_{i,j}$ with respect to the context $c_i$. For this, we first apply the same hard context retrieval method proposed by \citet{BERT-enhance} to filter out irrelevant utterances and to reduce the context size. Then, we put each context utterance paired with the response $r_{i,j}$ as the input to Topic-BERT to compute the corresponding topic vectors $\mT_j$ through the topic attention layer.    

We pass the topic vectors $\mT_j \in \mathbb{R}^{n\times d}$ through a scaled dot-product self-attention layer \cite{transformer} to learn all-pair topic relevance at the utterance level. Formally, 
\begin{equation}
\label{eq:Self-Attention}
\centering
     \mT_j^{'} = \softmax\big(\frac{ (\mT_j \mW_q) (\mT_j\mW_k)^{\top}}{\sqrt{d}}\big) (\mT_j\mW_v)
\end{equation}
where $\{\mW_q,\mW_k,\mW_v\} \in \mathbb{R}^{n\times d}$ are the query, key and value parameters, respectively, and $d$ denotes the hidden dimension of $768$. 

We add a max-pooling layer to select the most important information followed by a linear layer and a $\softmax$ to compute the relevance score of the response $r_{i,j}$ with the context $c_i$. Formally, 
\begin{equation}
\centering
    \hspace{-0.3em}f_{{\theta_{r}}}(u_k, r_{i,j}) = \softmax (\mW_r  (\text{maxpool}(\mT_j^{'})))
\end{equation}
where $\mW_r$ is the task-specific parameter. We use the standard cross entropy loss defined as:
\begin{equation}
\label{eq:response selection loss}
    \gL_{\text{rs}} = - \sum_{i,j} \mathbbm{1} (y_{i,j}) \log (f_{\theta_{r}}) 
\end{equation}
where $\mathbbm{1} (y_{i,j})$ is the one-hot encoding of the ground truth label.





\subsubsection{Topic Disentanglement}

For topic disentanglement (\cref{sec:taskform}), our goal is to find the ``reply-to'' links between the utterances (including the candidate response) to track which utterance is replying to which previous utterance.

For training on topic disentanglement, we simulate a sliding window over the entire (entangled) conversation. Each window constitutes a context $c_i = \{u_{1},u_{2}, \ldots, u_{n}\}$ and the model is trained to find the parent of $u_n$ in $c_i$, in other words, we try to find the reply-to link $(u_n, u_{n_{p}})$ for $1 \le {n_{p}} \le n$. 

For the input to our Topic-BERT (\Cref{fig:model}b), we treat $u_n$ as the response, thus allowing also response-response $(u_n, u_n)$ interactions through Topic-BERT's encoding layers to facilitate \emph{self-link} predictions (the fact that $u_n$ can point to itself).    


In the task-specific layer for disentanglement, we take the self-attended topic vectors $\mT_j^{'} = \{\vt_1^{'}, \ldots, \vt_n^{'}\}$ as input, and separate it into two parts: context topic vectors encapsulated in $\mT_c^{'} = \{\vt_1^{'}, \ldots, \vt_{n-1}^{'}\}$ $\in \mathbb{R}^{(n-1) \times d}$ and the response topic vector $\vt_n^{'}$ $\in \mathbb{R}^{d}$. In order to model high-order interactions between the response and context utterances, we compute the differences and element-wise products between them \cite{ESIM2}. We duplicate the response message $\vt_n^{'}$ to obtain $\mT_r^{'} \in \mathbb{R}^{(n-1) \times d}$ and concatenate them as:
\begin{equation}
    \mT^{''} = [\mT_r^{'},\mT_c^{'},\mT_r^{'} \odot \mT_c^{'}, \mT_r^{'} - \mT_c^{'}] \label{eq:esim}
\end{equation} 


Then, we compute the {\it reply-to} distribution as: $h_{\theta_{d}}(u_n, c_i) = \softmax (\mT{''} \vw_d) \in \mathbb{R}^{n \times 1}$, and optimize with the following cross-entropy loss:  
\begin{equation}
\label{eq:topic disentanglement loss}
    \gL_{\text{dis}} = -\sum_{j=1}^n \mathbbm{1} (y_{j}) \log (h_{\theta_{d}})
\end{equation}

\noindent For inference, we compute~$\argmax_j h_{\theta_{d}}(u_n, c_i)$.


\subsubsection{Multi-task Learning}

We jointly train the three tasks: response selection, topic prediction and topic disentanglement, which share the same topic attention weights to benefit each other. Response selection should benefit from dynamic topic prediction and disentanglement. Similarly, topic prediction and disentanglement should benefit from the response prediction. 
The overall loss is a combination of the three task losses from Equations \ref{eq:topic prediction loss}, \ref{eq:response selection loss}, and \ref{eq:topic disentanglement loss}:
\begin{equation}
    \gL = \alpha \gL_{\text{rs}} + \beta \gL_{\text{topic}} + \gamma \gL_{\text{dis}}
\end{equation}
where $\alpha$, $\beta$, and $\gamma$ are parameters which are chosen from $[0,0.1,0.2,...,1]$ by optimizing our model response selection accuracy on dev dataset.



\section{Experiments}
In this section, we present our experiments, including the datasets, experimental setup, evaluation metrics, and the results with analysis.

\paragraph{Datasets and Setup}

Considering multi-party conversations, we adopt a publicly available Ubuntu dataset from DSTC-8 track 2 ``NOESIS II: Predicting Responses"\cite{kim2019eighth}. This dataset consists of four tasks and we use the datasets from three of them, including disentangled conversations for response selection (Task 1), multi-party (mostly entangled) conversations for response selection  (Task 2), which is ideal for our main response selection evaluation, and a section of IRC channel with reply-to link annotations for conversation disentanglement (Task 4). \Cref{table:dstc8-data-1} shows some basic  statistics about the datasets. 

\begin{table}[t]
\centering
\small
\setlength{\tabcolsep}{3pt}
\scalebox{1}
{\begin{tabular}{cl|ccc}
\toprule
{\bf Tasks} &  & {\bf Train} & {\bf Validation} & {\bf Test}  \\
 \midrule
        & \# Dialog & 225,367 & 4,827 & 5,529 \\
 Task 1 & \# Avg. Turns & 6.0 & 6.1 & 6.1 \\
        & \# Avg. Speakers & 2.4 & 2.4 & 2.4 \\
\midrule
        & \# Dialog & 112,262 & 9,565 & 9,027 \\
 Task 2 & \# Avg. Turns & 54.7 & 54.3 & 54.6 \\
        & \# Avg. Speakers & 19.6 & 18.5 & 18.2 \\
\midrule
        & \# Link & 69,395 & 2,607 & 5,187 \\
 Task 4 & \# Avg. Turns & 3.9 & 3.6 & 3.0 \\
        & \# Avg. Speakers & 1.5 & 1.8 & 1.5 \\
\bottomrule
\end{tabular}}
\caption{DSTC-8 Ubuntu Dataset Statistics.}
\label{table:dstc8-data-1}
\end{table}

\paragraph{Evaluation Metrics}
DSTC-8 Track 2 considered a range of metrics for comparing models. We follow their evaluation metrics. Recall@N is used for topic prediction and response selection, which counts how often the correct answer is within the top N specified by the model. The values of N can be 1, 5, and 10. In addition, the Mean Reciprocal Rank (MRR) is also considered which is a widely used metric from the ranking literature. 

For disentanglement, we use precision, recall and F-scores (\wrt\ true-class) for link-level  predictions. Similarly, for topic prediction, we use precision, recall and F-scores for correctly classifying the actual response to be in the same topic.




\subsection{Experiment I: Response Selection}
\label{RS}
\paragraph{Baseline Models.} We compare the proposed Topic-BERT approach with several existing and state-of-the-art approaches for response selection:
\begin{itemize}[leftmargin=*,itemsep=0.1pt]
\item\textbf{BERT.} We adopt the vanilla pretrained uncased $\text{BERT}_{\text{base}}$\footnote{\url{https://github.com/huggingface/transformers}} as the base model, and follow \cite{gu2020pre} to post-train $\text{BERT}_{\text{base}}$ for 10 epochs on DSTC-Task 1 (response selection in a single-topic dialog). We take the whole context with the response as one input sequence. We then finetune it on Task 2's response selection for 10 more epochs. More details can be found in Appendix. 

\item\textbf{ToD-BERT.} This is a domain-specific pretrained BERT from \citet{wu2020tod}, which is pretrained on a combination of 9 Task-oriented Dialogue datasets and surpasses BERT in several downstream response selection tasks. 
\item\textbf{BERT-ESIM.} This model ensembles both ESIM \cite{ESIM} and BERT with gradient boosting classifier, and ranks the second best in DSTC-8 response selection \cite{BERT-ESIM}.
\item\textbf{Adapt-BERT.} This is based on BERT model with task-related pretraining and context modeling through hard and soft context modeling, and ranks as top-1 in the DSTC-8 response selection challenge \cite{BERT-enhance}.
\end{itemize}

\paragraph{Results.} From \Cref{table:response selection}, we can see that our Topic-BERT model outperforms the baselines by a large margin. By examining our model in detail, we found that our context filtering, self-supervised topic training and topic attention contribute positively to our model, boosting the metric of Recall@1 from 0.287 ($\text{BERT}_\text{base}$) to 0.696 (Topic-BERT with standalone response selection task). This shows our topic pretraining with task related data improves BERT for response selection task. 

Furthermore, the performance continues to increase from 0.696 to 0.710, when we jointly train response selection and topic prediction (2nd last row), validating an effective utilization of topic information in selecting response. 
Then we replace topic prediction with disentanglement, which further improves from 0.710 to 0.720, showing response selection can utilize topic tracing by sharing the connection of utterances. Finally, our Topic-BERT with the multi-task learning achieves the best result (0.726) and significantly outperform the prior state-of-the-art Adapt-BERT in DSTC-8 response selection task \cite{kim2019eighth}.

We further compute BLEU4 SacreBLEU \cite{post2018call} for the incorrectly selected responses by Topic-BERT and ToD-BERT. From \Cref{table:BLEU score}, we see that responses retrieved by Topic-BERT are more relevant even if they are not the top one.  

\begin{table}[t!]
\centering
\setlength{\tabcolsep}{3pt}
\scalebox{0.80}
{

\begin{tabular}{l|cccc}
\toprule
{\bf Model} & {\bf Recall@1} & {\bf Recall@5} & {\bf Recall@10} & {\bf MRR}  \\
\midrule
\midrule
$\text{BERT}_{\text{base}}$ & 0.287 & 0.503 & 0.572 & 0.351\\
BERT{\scriptsize +post-train} & 0.532 & 0.797 & 0.840 & 0.677\\
ToD-BERT & 0.588 & 0.823 & 0.885 & 0.691 \\
\midrule
Adapt-BERT & 0.706 & 0.916 & 0.957 & 0.799\\
\midrule
Topic-BERT & {\bf 0.726} & {\bf 0.930} & {\bf 0.970} & {\bf 0.807} \\
~~~\textminus TP & 0.720 & 0.927 & 0.964 & 0.803  \\
~~~\textminus D & 0.710 & 0.924 & 0.960 & 0.800 \\
~~~\textminus TP \textminus D & 0.696 & 0.910 & 0.950 & 0.790 \\

\bottomrule
\end{tabular}}

\caption{Response selection results on DSTC-8 Ubuntu. 
``-TP" means our model excluding topic prediction loss and ``-D" means excluding topic disentanglement loss. Adapt-BERT results are obtained from \cite{BERT-enhance}, other DSTC-8 released baselines are in Appendix.}
\label{table:response selection}
\end{table}

\begin{table}[t]
\centering
\setlength{\tabcolsep}{3pt}
\scalebox{0.8}
{\begin{tabular}{l|c|cccc}
\toprule
& &\multicolumn{4}{|c}{\textbf{Precision@N-gram}} \\
{\bf Model} & {\bf BLEU4} & {\bf N = 1 } & {\bf N = 2} & {\bf N = 3} & {\bf N = 4} \\
\midrule
ToD-BERT & 0.67 & 7.568 & 1.894 & 0.218 & 0.065\\
Topic-BERT & 0.75 & 7.876 & 2.032 & 0.250 & 0.078\\\hline
\end{tabular}}
\caption{BLEU4 and N-gram precision are calculated using SacreBLEU on incorrectly selected responses.}
\label{table:BLEU score}
\end{table}

\subsection{Experiment II: Topic Prediction}
\label{BERT TP}
This experiment aims to examine how significant our Topic-BERT can improve over the baselines on the topic prediction task, which is important for both response selection and topic disentanglement. 
\paragraph{Baseline Models.}
\begin{itemize}[leftmargin=*,itemsep=0.1pt]
\item\textbf{BERT.} {We use our post-trained $\text{BERT}_{\text{base}}$ from \cref{RS} and fine-tune it on Task 1 topic sentence pairs as our BERT baseline for topic prediction.} 
\item\textbf{ToD-BERT.} We adopt our post-trained ToD-BERT and fine-tune it with our obtained topic sentences pairs 
as the ToD-BERT baseline. 
\end{itemize}

\paragraph{Results.} \Cref{table:topic prediction} gives the topic prediction results on DSTC-8 {task-1}. From the results, we can see that our Topic-BERT outperforms the baselines BERT and ToD-BERT significantly in the topic prediction task. Compared with our pretrained Topic-BERT without fine-tuning (last row), the proposed topic attention further enhances the topic matching of two utterances by improving the F-score by 1.5\% (from 0.813 to 0.828). Joint training with response selection or disentanglement tasks show similar effect on topic prediction tasks, and the contextual topic information sharing by Topic-BERT multi-task model add a marginal improvement in topic prediction. Compared with vanilla BERT, ToD-BERT \cite{wu2020tod} makes substantial improvement for the topic prediction task, but not as significant as ours. This further confirms the importance and efficacy of our learning scheme. {Meanwhile, if we compare our pretrained Topic-BERT without fine-tuning (last row) with the BERT model that does not use STP (first row), the significant improvement gives us an impression on how much our Topic-BERT benefits from the STP loss.}
\begin{table}[t!]
\centering
\scalebox{0.8}
{\begin{tabular}{l|ccc}
\toprule
{\bf Model} & \quad{\bf Precision} & {\bf Recall} & {\bf F-Score} \\
\midrule
\hline
BERT & 0.523 & 0.482 & 0.502 \\
ToD-BERT & 0.626 & 0.563 & 0.593\\
\midrule
Topic-BERT &  0.890 & {\bf 0.847} & {\bf 0.868} \\
~~~\textminus D & {\bf0.891} & 0.845 & 0.867 \\
~~~\textminus RS &  0.889 & 0.840 & 0.864 \\
~~~\textminus D \textminus RS & 0.866 & 0.793 & 0.828  \\
\; w/o FT & 0.848 & 0.781 & 0.813 \\
\bottomrule
\end{tabular}}
\caption{Topic prediction results on DSTC-8 Ubuntu. 
``w/o FT" means our Topic-BERT without fine-tuning, ``-RS" means our model excluding the Response Selection loss, ``-D" means excluding Disentanglement loss.}
\label{table:topic prediction}
\end{table}

\subsection{Experiment III: Disentanglement}
This experiment aims to examine how well can Topic-BERT tackle the topic disentanglement task. 

\begin{table}[t]
\centering
\setlength{\tabcolsep}{3pt}
\scalebox{0.8}
{\begin{tabular}{l|ccc}
\toprule
{\bf Model} & \; {\bf Precision} & \;\;\quad{\bf Recall}\quad\;\;\quad & {\bf F-Score}   \\
\midrule
\hline
BERT & 0.431 & 0.417 & 0.424  \\
MH BERT & 0.539 & 0.517 & 0.528\\
ToD-BERT & 0.612 & 0.603 & 0.607\\
Feed-Forward & 0.748 & 0.718 & 0.733 \\
\midrule
Topic-BERT & {\bf 0.754} & {\bf 0.725} & {\bf 0.739}\\
~~~\textminus TP & 0.749 & 0.727 & 0.737  \\
~~~\textminus RS & 0.705 & 0.692 & 0.698 \\
~~~\textminus TP \textminus RS& 0.689 & 0.678 & 0.683  \\
\bottomrule
\end{tabular}}
\caption{Disentanglement results on DSTC-8 Ubuntu. 
``-RS" means our model excluding Response Selection loss, and ``-TP" means excluding Topic Prediction loss. 
}
\label{table:disentanglement}
\end{table}

\paragraph{Baseline Models.}
\begin{itemize}[leftmargin=*,itemsep=0.1pt]

\item\textbf{BERT} \& \textbf{ToD-BERT}
We use our fine-tuned {BERT} and {ToD-BERT} models in \cref{BERT TP} as our baselines by taking the history of utterances ($u_1, \ldots,  u_{n-1}, u_n$) and pair each with the current utterance $u_n$ itself from a dialogue as input. Following \cite{gu2020pre}, A single-layer BiLSTM is applied to extract the cross message semantics of \texttt{[CLS]} outputs. Then we take the differences and element-wise products (Eq. \ref{eq:esim}) between the history and current utterance. Finally, a feedforward layer is used for link prediction. 




\item\textbf{Feed-Forward.} This is the baseline model\footnote{\url{https://github.com/dstc8-track2/NOESIS-II/tree/master/subtask4}} from DSTC-8 task organizers that has the best result for task 4 \cite{acl19disentangle}, which is trained by employing a two-layer feed-forward neural network on a set of {\it 77 hand engineered features} combined with word average embeddings from pretrained Glove embeddings.
\item\textbf{Masked Hierarchical (MH) BERT.} This is a two-stage BERT proposed by \citet{zhu2019did} to model the conversation structure, in which the low-level BERT is to capture the utterance-level contextual representation between utterances, and the high-level BERT is to model the conversation structure with an ancestor masking approach to avoid irrelevant connections. 
\end{itemize}
\paragraph{Results.}
From the results in \Cref{table:disentanglement}, we can see that
our Topic-BERT achieves the best result and outperforms all the BERT based baselines significantly. This shows our multi-task learning can enrich the link relationship for improving disentanglement together with topic prediction and response selection. The improvement of Topic-BERT over the baseline model using feed-forward network and hand-crafted features is relatively less, but our approach is able to avoid manual feature engineering. {Many of these features are dataset/domain specific and they do not generalize across datasets/domains.}


\subsection{Experiment IV: Evaluation on New Task}
\begin{table}[t!]
\centering
\setlength{\tabcolsep}{3pt}
\scalebox{0.7}
{\begin{tabular}{l|ccc}
\toprule
{\bf Model} & \quad\textbf{Recall}{10}@1\quad & \quad\textbf{Recall}{10}@2\quad  & \quad\textbf{Recall}{10}@5\quad\\
\midrule
\hline
DL2R & 0.626 & 0.783 & 0.944 \\
Multi View & 0.662 & 0.801 & 0.951\\
$\text{SMN}_{\text{dynamic}}$ & 0.726 & 0.847 & 0.961 \\
AK-DE-biGRU & 0.747& 0.868 & 0.972 \\
DUA & 0.752 & 0.868 & 0.962 \\
DAM & 0.767 & 0.874 & 0.969 \\
IMN & 0.777 & 0.888 & 0.974 \\
ESIM & 0.796 & 0.894 & 0.975\\
$\text{MRFN}_{\text{FLS}}$ & 0.786 & 0.886 & 0.976\\
\midrule
\midrule
$\text{BERT}_{\text{base}}$ & 0.817 & 0.904 & 0.977 \\
BERT-DPT & 0.851 & 0.924 & 0.984 \\
Topic-BERT 
& {\bf0.861} & {\bf0.933} & {\bf0.985}\\
\bottomrule
\end{tabular}}
\caption{Response selection results on Ubuntu Corpus v1. 
All other results are from \cite{whang2019domain}.
}
\label{table:ubuntu v1response selection}
\end{table}
Finally, we examine our Topic-BERT's transferability on a new task based on another Ubuntu Corpus v1 dataset by comparing with various state-of-the-art response selection methods in \Cref{table:ubuntu v1response selection}. Ubuntu Corpus V1 contains 1M train set, 500K validation and 500K test set \cite{lowe2015ubuntu}.

\paragraph{Baseline Models.} Here we mainly introduce the state-of-the-art baseline: BERT-DPT \cite{whang2019domain}, which fine-tunes BERT by optimizing the domain post-training (DPT) loss comprising both NSP and MLM objectives for response selection. 
Details of other baselines can be found in Appendix.  
\paragraph{Results.} Our Topic-BERT with standalone response selection task fine-tuned on Ubuntu Corpus v1 outperforms the state-of-the-art BERT-DPT, 
improved by about $1\%$ for $\text{Recall}_{10}$@1.
This result shows that the learned topic relevance in Topic-BERT can be potentially transferable to a novel task, the topic information influences the response selection positively, and our utterance-level topic tracking is effective for response selection.


\section{Conclusion}
This paper presented a new formulation of response selection in multi-party conversations from a novel dynamic topic tracking perspective. 
Based on our new formulation, we propose Topic-BERT for response selection in multi-party conversations, which consists of two steps: (1) a topic-based pretraining to embed topic information into BERT with self-supervised learning, and (2) a multi-task learning on our pretrained model by jointly training response selection and dynamic topic prediction and disentanglement tasks. Empirically the proposed Topic-BERT achieved the state-of-the-art results on the DSTC8 Ubuntu IRC datasets.

\bibliography{anthology,emnlp2020}
\bibliographystyle{acl_natbib}

\end{document}


\maketitle


\section{Appendix A: Details of Experiments}

\subsection{Details of Datasets}
Consider the multi-party conversations, we adopt a publicly available Ubuntu dataset from DSTC-8 track of ``NOESIS II: Predicting Responses"\cite{kim2019eighth}. This Ubuntu dataset consists of four tasks and we use datasets from three of them. Task 1 is a single-topic multi-party dialogue setting which is considered for our topic prediction, and each test sample of the test set consists of 1 positive response and 4 negative responses randomly selected from 99 negatives. Task 2 is a long Ubuntu chat log with multi-party conversations on-going simultaneously which is ideal for our response selection evaluation, and each test sample in the test set contains 1 positive response and 99 negative responses. Task 4 is formed by multi-party conversations with link annotations, where each instance has 1500 utterances with around 500 annotation links. The numbers of conversations in task 4 datasets are: 17,751 in the training set, 785 in the dev set, and 964 in the test set. The corresponding numbers of links are: 69,395 in the training set, 2,607 in the dev set, and 5,187 in the test set. Table 1 in the main paper gives the statistics of this dataset used in our experiments. 

\subsection{Evaluation Metrics}
Recall@N is used for topic prediction and response selection, which counts how often the correct answer is within the top N specified by the model. The values of N can be 1, 5, and 10. In addition, the Mean Reciprocal Rank (MRR) is also considered which is a widely used metric from the ranking literature. 

\begin{table}[ht]
\centering
\small
\setlength{\tabcolsep}{3pt}
\scalebox{1}
{\begin{tabular}{c|ccc}
\toprule
{\bf Tasks@Track2} & {\bf Turns (Avg)} & {\bf Speakers (Avg)}  \\
 \midrule
 \hline
 Task 1 Train & 6.0 & 2.4 \\
 Task 1 Val & 6.1 & 2.4 \\
 Task 1 Test & 6.1 & 2.4 \\
 \hline
 Task 2 Train & 54.7 & 19.6  \\
 Task 2 Val & 54.3 & 18.5  \\
 Task 2 Test & 54.6 & 18.2  \\
 \hline
 Task 4 Train & 3.9 & 1.5  \\
 Task 4 Val & 3.6 & 1.8  \\
 Task 4 Test & 3.0 & 1.5  \\
\bottomrule

\end{tabular}}
\caption{DSTC-8 Ubuntu Dataset Statistics: Average Turns and Average Speakers per conversation}
\label{table:dstc8-data}
\vspace{-0.11in}
\end{table}

\subsection{Implementation Details}
\label{BERT topic pairs}
We adopt the vanilla pretrained uncased $\text{BERT}_{\text{base}}$ (which has 12 layers with a hidden size of 768, provided by hugging face library) \footnote{\url{https://github.com/huggingface/transformers}} to initialise Topic-BERT with uncased $\text{BERT}_{\text{base}}$ configuration. In addition, according to section of Topic-BERT Pretraining, we obtained 13561221 train sentence pairs, 576414 validation sentence pairs and 435680 test sentence pairs from Task 1 corresponding dataset. Inspired by \cite{gu2020pre,BERT-enhance}, we posttrained our input-encoder (uncased $\text{BERT}_{\text{base}}$) 10 epochs more with the training topic sentence pairs and the batch size is set to 96, the MLM training objective was on positive training pairs only. Then 2 more epochs with validation topic sentence pairs was used for fine-tuning.  

We used cross-entropy loss and Adam optimizer with a learning rate of 2e-5 and a linear decay schedule with warm-up 0.1. Regarding some hyper-parameters, the maximum sequence length of the concatenation of a context-response pair was set to 192 (96 for response), and the ratio of dropout is 0.1 to prevent over-fitting. Then our trained Topic-BERT will be used to encode input sentences pairs (10 utterance pairs for each batch to limit the context) to obtain the topic vectors for the end 3 tasks. 4 Tesla V100-SXM2 GPUs was used to train Topic-BERT, the training time for each epoch is around 30.2 hours and the validation time is around 45.2 minutes to complete one epoch for posttraining.

\section{Appendix B: Details of Baseline Models}
\begin{table}[t!]
\centering
\setlength{\tabcolsep}{3pt}
\scalebox{0.85}
{\begin{tabular}{l|cccc}
\toprule
{\bf Model} & {\bf Recall@1} & {\bf Recall@5} & {\bf Recall@10} & {\bf MRR}  \\
\midrule
\hline

Switch-BERT & 0.506 & 0.755 & 0.834 & 0.621\\
BERT-ESIM & 0.596 & 0.847 & 0.904 & 0.707 \\
XLNet & 0.637 & 0.856 & 0.914 & 0.735 \\
Adapt-BERT & 0.706 & 0.916 & 0.957 & 0.799\\\hline

\end{tabular}}
\vspace{-0.5em}
\caption{Released response selection results on DSTC-8 Ubuntu task 2. The Switch-BERT result is obtained from \cite{gu2020pre}, BERT-ESIM result is obtained from \cite{BERT-ESIM}, Adapt-BERT result is obtained from \cite{BERT-enhance}, XLNet result is obtained from \cite{BERT-SM}}
\label{table:response selection base}
\vspace{-0.5em}
\end{table}

%
%
%
%
\subsection{Baselines for Response Selection}
\label{BRS}
\paragraph{Baseline Models.} We compare the proposed Topic-BERT approach with several existing and state-of-the-art approaches for response selection:
\begin{itemize}[leftmargin=*,itemsep=0.1pt]
\item\textbf{BERT.} The vanilla pretrained uncased $\text{BERT}_{\text{base}}$ \footnote{\url{https://github.com/huggingface/transformers}} was the base model. We follow \cite{BERT-enhance} to posttrain $\text{BERT}_{\text{base}}$ on task 1 dataset 10 more  epochs with NSP and MLM objectives. The maximum sequence length was set to 512 for the document-level input with a concatenation of the whole context and response to capture the most present tokens. Adam optimizer with a learning rate of 2e-5 and a linear decay schedule with warm-up 0.1. Batch size is 16.

In fine-tuning, 10 more epoch was used for response selection on task 2 dataset. Adam optimizer with a learning rate of 3e-5 and a linear decay schedule with warm-up 0.06 was selected. For response selection, the candidate pool may not contain the correct response, so we need to choose a threshold. When the probability of positive labels was smaller than the threshold, we predicted that candidate pool did not contain the correct response. The threshold was selected from the range [0.7, 0.75, .., 0.95]\cite{gu2020pre}, based on the validation set, it was set to 0.90 finally. We used the validation set to set the stop condition to select the best model for testing.   

\item\textbf{ToD-BERT.} ToD-BERT \cite{wu2020tod} is a pre-trained language model with 9 task-oriented dialogue datasets, it surpasses BERT in downstream response selection task. We compare our topic pretrained BERT with the ToD-BERT. We initialized BERT model with ToD-BERT, and posttrain it 10 more epochs on task 1 dataset with NSP and MLM objectives. A limit of maximum sequence is set to 512 by taking a document-level context history and response as input sequence. Adam optimizer with a learning rate of 1e-5 and a linear decay schedule with warm-up 0.1.

There are 10 more epochs for fine-tuning on task 2 response selection as well. Adam optimizer with a learning rate of 2e-5 and a linear decay schedule without warm-up was selected. Based on the validation set, the none reply threshold was set to 0.95. 

\item\textbf{Switch-BERT.} \citet{gu2020pre} introduced switch embedding to indicate a switch of turn when the conversation is proceeding. External data (Ubuntu FQA) will be utilized for BERT pretraining. We list down their reported result as baseline in \cref{table:response selection base}.

\item\textbf{BERT-ESIM.} Model ensembling of ESIM and BERT with gradient boosting classifier which rank as the second team in DSTC-8 response selection \cite{BERT-ESIM}. ESIM is the state-of-the-art model in DSTC-7 by applying cross attention to get the inter-sentence matching. Then ESIM and BERT will be ensembled with a gradient boost classifier. We list down their reported result as baseline in \cref{table:response selection base}. 

\item\textbf{Adapt-BERT.} BERT model with task-related pretraining and context modeling through hard and soft context modeling which rank as the top-1 team in DSTC-8 response selection challenge \cite{BERT-enhance}. The hard context retrieval and data augmentation make the response selection performance more accurate. We list down their reported result as baseline in \cref{table:response selection base}. 
\end{itemize}

\subsection{Baselines for Topic Prediction}
\label{BERT TP}
\begin{itemize}[leftmargin=*,itemsep=0.1pt]
\item\textbf{BERT.} \cref{BRS} posttrained BERT will be fine-tuned with topic sentence pairs which are generated from task 1 for 10 epochs and the batch size is set to 96. We used cross-entropy loss and Adam optimizer with a learning rate of 2e-5 and a linear decay schedule with warm-up 0.1. Regarding some hyper-parameters, the maximum sequence length of the concatenation of a context-response pair was set to 192 (96 for response), and the ratio of dropout is 0.1 to prevent over-fitting.
\item\textbf{ToD-BERT.} The posttrained ToD-BERT in \cref{BRS} will be fine-tuned with 10 more epochs on task 1 topic sentence pairs with a batch size of 96 and up to 10 utterance counts per batch, the maximum sequence length is 192 (96 for response), an Adam optimizer with a learning rate of 1e-5 and a linear decay schedule with warm-up 0.1 were used in this experiment.
\end{itemize}

\subsection{Baselines for Topic Disentanglement}

\begin{itemize}[leftmargin=*,itemsep=0.1pt]
\item\textbf{BERT and ToD-BERT.}
We use our fine-tuned \textbf{BERT} and \textbf{ToD-BERT} model mentioned in \cref{BERT TP} as our baselines by taking the context utterance-response pairs from a dialogue as inputs, following \citet{gu2020pre} the initial learning rate is 2e-5 with maximum sequence length of 100, a single-layer BiLSTM with hidden size of 384 (BiLSTM output hidden size will be 768 to match with BERT output hidden size) is employed on top of the outputs of BERT [cls] outputs to capture the semantics across different messages and a single-layer feedforward NN with 768×4 hidden units will be the classifier to calculate the link-relevance score. The classifier has 3072 hidden units with 0.1 dropout to prevent overfitting. An Adam optimizer with a learning rate of 2.5e-5 and a linear decay schedule with warm-up 0.05 were used in this experiment.

\item\textbf{Feed-Forward.} We use the feed-forward model from \citet{acl19disentangle} as the baseline model. In DSTC-8, \citet{acl19disentangle} provided a fully trained model. \footnote{\url{https://github.com/dstc8-track2/NOESIS-II/tree/master/subtask4}} The provided model is trained using a two-layer neural network. The input is 77 hand engineered features combined with word average embeddings from pretrained Glove embeddings. We list down their reported result as baseline in main paper Table 4 to compare with.

\item\textbf{Masked Hierarchical (MH) BERT.} It is a two stage BERT to model the conversation structure where the low level BERT is to capture the utterance-level contextual representation between utterances. And the high level BERT is to modeling the conversation structure with an ancestor masking approach to avoid irrelevant connections \cite{zhu2019did}. We list down their reported result as baseline in main paper Table 4 to compare with.
\end{itemize}

\subsection{Baselines for Experiment IV}
The Topic-BERT with standalone response selection setting is fine-tuned with Ubuntu Corpus V1 with following hyperparameter setting, Adam optimizer with initial learning rate 5e-5, the maximum sequence length of the concatenation of a context-response pair was set to 192 (96 for response),batch size is 96 and 10 utterance pairs per batch, 10 epochs in total. 

Below are the most commnon baseline models for Ubuntu Corpus V1 with a brief introduction. We list down their reported result as baseline in main paper Table 5 to compare with \cite{whang2019domain}.

\begin{itemize}[leftmargin=*,itemsep=0.1pt]
\item\textbf{DL2R.} It is a deep neural network to capture the relations cross sentence pairs. Query will be reformulated along with context utterance to enrich the contextual information for response selection \cite{yan2016learning}.

\item\textbf{Multi View.} It is a combination of word sequence model and utterance sequence model, the word-view and utterance-view will be used through a hierarchical RNN for response matching \cite{zhou2016multi}.

\item\textbf{SMN.} Response will be matched with each contextual utterances at multiple levels of granularity, then SMN will accumulate these matching information to select response \cite{SMN}.

\item\textbf{AK-DE-biGRU.} It is attention based dual encoder with external data to incorporate domain knowledge to improve response selection \cite{chaudhuri2018improving}

\item\textbf{DUA.} \citet{zhang2018modeling} proposed utterance aggregation approach with attention matching for response selection.

\item\textbf{DAM.} It is a transformer based model to utilize utterances self-attention and context-to-response cross attention to leverage the hidden representation at multi-grained level \cite{DAM}.

\item\textbf{IMN \& ESIM} Both of them enrich sentence representation with inter-sentence matching to solve response selection problem \cite{gu2019interactive,ESIM2,ESIM1}. 
\end{itemize}
\begin{table*}[ht!]
\centering
\setlength{\tabcolsep}{3pt}
\scalebox{0.8}
{\begin{tabular}{c|c}
\toprule
\textbf{Speaker}& \textbf{Utterance}  \\
 \midrule
 None & None \\
 None & None \\
 participant\_23 & <participant_0> participant_1, ctrl+alt+del? \\
 participant\_25 & i would just like to know, how can i found out the support webcams for dapper? \\
 participant\_23 & in sessions what do i have to add to the start of my command to run the command as root when i boot up ? \\
 participant\_25 & *supported \\
 participant\_23 & \textbf{to start lampp} \\
 participant\_23 & \textbf{opt/lampp/lampp start} \\
 participant\_25 & is there a command in shell, or a section in the ubuntu website to find out all the supported webcams for dapper drake? \\
 participant\_23 & ok ty participant\_20\\
\midrule
ToD-BERT & participant\_25 , permission on /home/luigi/wwwebserver http://pastebin.ubuntu.com/75435/ \\
Topic-BERT & \textbf{ill put it in rc.local}\\
Ground-truth & ill put it in rc.local\\
 \bottomrule
\end{tabular}}
\caption{Top-1 response selection performance comparison}
\label{table:example1}
\end{table*}

\begin{table*}[ht!]
\centering
\setlength{\tabcolsep}{3pt}
\scalebox{0.8}
{\begin{tabular}{c|c}
\toprule
\textbf{Speaker}& \textbf{Utterance}  \\
 \midrule
 \hline
 participant\_0 & participant\_1, ctrl+alt+del? \\
 None & None \\
 participant\_1 & \bf participant\_0: no response, this is when i logged in \\
participant\_2 & \bf is medibuntu down? \\
 participant\_3 & I'm unable to browse a samba share on my own computer using nautilus. If I try to access a windows  \\
           & share I only get a empty window. Any hints/ideas?\\
 participant\_4 & hi, when I downlad files from the repositories, where in Ubuntu Linux they will be saved? \\
 participant\_0 & participant\_2, yes \\
 participant\_5 & participant\_4: depends on what you want. \\
 participant\_2 & \br participant\_0: is there a mirror somewhere? \\
 participant\_0 & participant\_1, power switch it is then\\
 
\midrule
\hline
ToD-BERT & participant\_5:  I want to save some of them to a usbstick after they where saved on my harddisk \\
Topic-BERT & \bf participant\_0: seems like even DNS is down\\
Ground-truth & participant\_0: seems like even DNS is down\\
\bottomrule
\end{tabular}}
\caption{Top-1 response selection performance comparison}
\label{table:example2}
\end{table*}
\section{Appendix C: Qualitative Evaluation}

\Cref{table:example1}, \Cref{table:example2} shows our qualitative comparison. Our topic pretraining strategy could retrieve response based on topic relevance, those informative non-topic relevant utterances will not influence our response selection. During the evaluation, we find that several response candidates such as "Yes",  "Haha", "np" may misleading the response selection. Those terms doesn't have intrinsic/specific meaning in topic, they could be over attended as the dominant factor for response selection. This also address our interest in future potential research area such as topic noise decoupling.





\bibliography{anthology,emnlp2020}
\bibliographystyle{acl_natbib}